# Developing a novel approach for Periapical dental radiographs segmentation

Elaheh Hatamimajoumerd
School of Electrical and Computer Engineering
Shiraz University
Shiraz, Iran
eh.hatami@gmail.com

Farshad Tajeripour
School of Electrical and Computer Engineering
Shiraz University
Shiraz, Iran
tajeri@shirazu.ac.ir

*Abstract*— **Image processing techniques has been widely used in dental researches such as human identification and forensic dentistry, teeth numbering, dental carries detection and periodontal disease analysis. One of the most challenging parts in dental imaging is teeth segmentation and how to separate them from each other. In this paper, an automated method for teeth segmentation of Periapical dental x-ray images which contain at least one root-canalled tooth is proposed. The result of this approach can be used as an initial step in bone lesion detection. The proposed algorithm is made of two stages. The first stage is pre-processing. The second and main part of this algorithm calculated rotation degree and uses the integral projection method for tooth isolation. Experimental results show that this algorithm is robust and achieves high accuracy.**

*dental image processing; bone lesion detection; tooth segmentation; root apex detection; periapical image*

## I. Introduction

Periodontal disease, are serious infections, that left untreated can lead to tooth loss. The word periodontal literally means "around the tooth". Periodontal disease is a chronic bacterial infection disease that affects the gums and bones supporting the teeth and leads to bone resorption. It can affect one tooth or many teeth. The bone resorption occurred mostly at the bone tissue close to the root apex. [1]

X-rays are a form of radiation like light or radio waves and pass through most objects, including the body. Once it is carefully aimed at the part of the body being examined, an x-ray machine produces a small burst of radiation that passes through the body, recording an image on photographic film or a special detector.

Different parts of the body absorb the x-rays in varying degrees. Dense bone absorbs much of the radiation while soft tissue, such as muscle, fat and organs, allow more of the x-rays to pass through them. As a result, bones appear white on the x-ray, soft tissue shows up in shades of gray and air appears black.

Until recently, x-ray images were maintained as hard film copy (much like a photographic negative). Today, most images are digital files that are stored electronically. These stored images are easily accessible and are frequently compared to current x-ray images for diagnosis and disease management [2].

Dentists utilize dental x-ray images in order to diagnose dental carries, bone resorption and, check the health of the tooth root and bone surrounding the tooth, check the status of developing teeth, and monitor the general health of teeth and jawbone.

Dental x-ray images are classified according to the view they are captured from, and their coverage. The most common type of X-ray taken is bitewing and Periapical x-ray images which are intraoral radiographic images. Fig. 1 shows these two types of intraoral radiograph. Bitewing X-rays show details of the upper and lower teeth in one area of the mouth. Each bite-wing shows a tooth from its crown to about the level of the supporting bone. Bite-wing X-rays are used to detect decay between teeth and changes in bone density caused by gum disease. They are also useful in determining the proper fit of a crown (or cast restoration) and the marginal integrity of fillings.

Periapical X-rays show the whole tooth, from the crown to beyond the end of the root to where the tooth is anchored in the jaw. Each Periapical X-ray shows this full tooth dimension and includes all the teeth in one portion of either the upper or lower jaw. Periapical X-rays are used to detect any abnormalities of the root structure and surrounding bone structure. It shows the entire tooth all the way down to the root tip. It can be used to see infection on the root tip of a tooth which cannot be seen clearly on any other type of x-ray. A Periapical x-ray can also give us a good view of bone loss around a tooth [3].

Figure 1.　　　(a) Bitewing X-ray. (b) Periapical X-ray

One of the most challenging parts in dental imaging is teeth segmentation and how to separate them from each other. Tooth segmentation is basic part of human identification using dental images. If we do the segmentation in a good way we can better detect the abnormality in tooth structures and diagnose Periapical bone lesion. There are many factors that affect the image quality. No imaging method is noise free but we must do our best to reduce image noise, so pre-processing is inevitable.

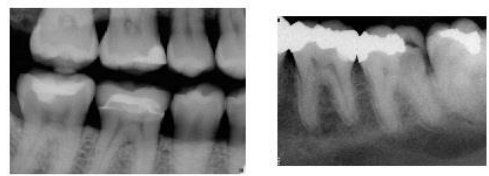

The first stage in tooth segmentation is estimating boundaries between every tooth and its neighbors. To fulfill this requirement, the segmentation algorithms apply the integral projection to find the gap between two adjacent teeth.

The paper is organizes as fallow. In section 2 we briefly talk about related work. We introduce proposed method for tooth segmentation in section 3. Experimental results are presented in section 4 and finally section 5 concludes the paper.

## II. RELATED WORK

Teeth segmentation methods involve variety types of algorithms. O. Nomir and M. Abdel-Mottaleb [4] used adaptive thresholding in order to separate teeth from the background. After that, they applied integral projection on images for tooth isolation. V. Phong-Dinh and L. Bac-Hoai [5] presented a segmentation method based on tooth anatomy. They extracted an equation that calculates the numerical certainty in order to decide which valleys in integral projection histogram are real jaw gaps. B. Vijayakumari et al [6] used Butterworth and Homomorphic filtering as a pre-process and applied integral projection. They finalize their segmentation by using canny edge detection to estimate the contour and shape of tooth in bite-wing intraoral radiograph. Separating teeth by vertical line parallel to y-axis don't work for all dental x-ray images. Depending on how radiograph images are taken and teeth structure, there is an angle between teeth and y-axis. [4] rotated the image in a small range of angles, e.g. [-20:20], calculating the vertical integral projection for each angle in a range, select separating lines and corresponding angles which produce the minimum vertical projection among all the rotation angles. For those images where the gap between the upper and lower jaws isn't exactly parallel to x-axis, Anil K Jain and Hong Chen [7] divided the bite-wing dental x-ray images into vertical strips and calculated the row which had minimum integral intensity in each strip then utilized B-Spline function to form a smooth curve which separates the upper and lower teeth. The use of B-Spline as occlusal curve is not good solution, because the B-Spline curve is sensitive to teeth alignment. Besides, the automation criterion is not satisfied [5]. After summing the intensity of pixels in each line perpendiculars to the smooth curve, lines which have minimum intensity are selected as teeth gaps. Zhou and Mottaleb [8] developed a fully automated system using jaw splitting. They segmented teeth by applying morphological filter to raw images before sparking Gradient Vector Flow snake. They also used snake method for tooth isolation.

These approaches tried to develop a segmentation method as an initial step for human identification system. Most of segmentation algorithm was done on bite-wing images and assumed that tooth can be separated from each other by vertical line parallel to y-axis. But in most of cases straight vertical line is not a good separator and we need to calculate the angles between teeth and axis. Here, we develop an automated segmentation method that uses property of root-canalled teeth in Periapical radiographs for calculating rotation degree and estimates the boundaries of teeth. Proposed method can be used as an initial step in bone lesion detection.

## III. PROPOSED METHOD

### A. pre-processing

Medical x-ray images are normally dark images with low visibility and require the enhancement of relevant details and suppression of unwanted artifacts, such as noise and other distortion. Reconstruction methods are trying to meet these contradictory requirements. The effects that contribute mostly to the degradation of image are: (1) the systematic non-homogeneous illumination caused by the x-ray source and its collimation. (2) The psychophysical contrast adjustment by the human visual system for the human eye according to the highlight levels of the surrounding (background) illumination. (3) The impact of the photon scatters noise statistics. [9]

In order to improve both contrast and intensity illumination evenness simultaneously, this algorithm uses an image enhancement method that utilize frequency domain filters like Butterworth and Homomorphic filters [6]. Applying Homomorphic filtering, high frequency components are increased and low frequency components are decreased, because high frequency component are assumed to represent mostly the illumination in the scene, that is high pass filtering is used to surpass low frequency and amplify high frequency in the log intensity domain [10]. For better performance, we have done Butterworth filter on input images to obtain uniform sensitivity to all frequencies. After that, we use obtained images as inputs to Homomorphic filter. An n-order Butterworth filter is given in the term of transfer function H(s) as:

$$G^2(\omega) = |H(j\omega)|^2 = \frac{G_0^2}{1+\left(\frac{\omega}{\omega_c}\right)^{2n}} \quad (1)$$

Figure 2.　　(a) original image. (b) Butterworth filtered (c).

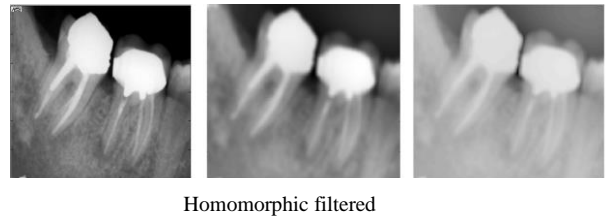

Homomorphic filtered

As what was done in [6], we choose the order of this filter to be 2. Fig.2 shows result of pre-processing methods. Fig.2 (a) illustrates original image, fig.2 (b) is the output of Butterworth filter and fig.2(c) is the result of Homomorphic filter applied on Butterworth filtered image.

### B. Integral projection

We can classify the area in the dental x-ray image into 3 groups: teeth, gum, and air. A tooth maps to an area with mostly "bright" gray scale, while gums map to area with "mid-range" gray scales, and air maps to dark gray scales [9]. We can benefit from these properties in tooth segmentation.

At first we assume that we can split each tooth from its neighbors by vertical line paralleled to y-axis. We consider a simple case of integral projection in which pixels are projected orthogonally onto x-axis. $proj(x)$ the accumulative intensity of pixels in each column of image function $f(x,y)$ $0 \leq x < w-1; 0 \leq y \leq h-1$; in which $w$ $and$ $h$ are the number of pixels in each row and column of image. series $\{proj(x_0), proj(x_{01}), ..., proj(x_{w-1})\}$ forms a graph of integral intensity. The projection function is defined as

(2) $\quad proj(y) = \sum_{x=0}^{w} f(x,y)$

In the other word, we sum the intensities of pixels located in each line parallel to y-axis. Since the gaps between teeth are darker than teeth area, $proj(x)$ in these lines have less value than other area and forms valleys in integral projection histogram. Due to internal noise, number of valleys often is more than gaps, and valley are candidate for gaps. We present a solution to overcome this situation.

We try to reduce the internal noise by smoothing filters. We apply smoothing filters like mean mask, Gaussian and wiener filter. We decrease the effect of sudden change in gray level intensity by applying these kinds of filters. Fig3 (a) shows the integral projection histogram before smoothing filters, while fig 3(b) illustrates the integral projection after using smoothing filters.

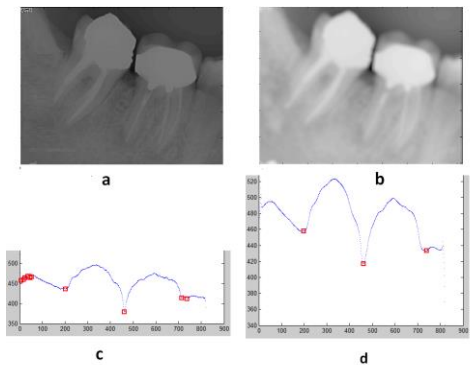

Figure 3. (a).pre-processed image without using smoothing filter (b). Pre-processed image using smoothing filters (c). Integral projection of image (a) (d). Integral projection of image (b)

So we modified pre-processing step. At the beginning we use mean mask for smoothing in which the squared mask is moved over the dental x-ray images. The central pixel intensity is replaced with average value by calculating average intensity of pixels mask covers them. We do this for all pixels on image. After that wiener filter was done on mean images. We apply a 2-D adaptive noise removal filtering which uses a pixel-wise adaptive wiener method base on statistical estimation from a local neighborhood of size m-by-n pixels to estimate local image mean and standard deviation. Finally we pass the obtained image through Gaussian filter.

We also assume that difference between gaps must be more than a threshold; if the difference between two valleys is fewer than threshold, we select the valley which has less value in integral projection histogram.

Fig.4. represents some of the images in our dataset. As it is shown, in most of the cases, we can't split teeth with a straight vertical line and we need to rotate the vertical line or image.

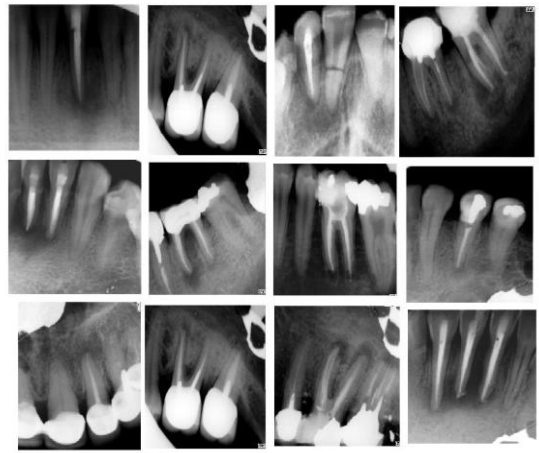

Figure 4. Sample of Periapical X-ray images from dataset

### 1) Estimation of rotation degree

Different parts of the body absorb the x-rays in varying degrees. Dense bone absorbs much of the radiation while soft tissue, such as muscle, fat and organs, allow more of the x-rays to pass through them. As a result, bones appear white on the x-ray, soft tissue shows up in shades of gray and air appears black. A root canal removes the pulp which is soft tissue inside the tooth and replaces it with filling material that appears bright like dentin area in X-ray images. By knowing this information, we divided pre-processed image into three parts. (The height of image is divided by tree), and we select the mid part, so we are sure that filled area of root-canalled tooth or teeth exists in this part.

We scan the first row in this part and plot the row intensity histogram. We store the position of each maximum point in different sets. For each row from the next to the end of selected part, after finding the maximum points, we calculate distance of every maximum point from all the points recently added to the sets. We assign the position of each maximum point in this row to the set contains nearest point from previous row. Each set can represent a part of pulp area that was filled with medical material, in the other hand; points in each set trace a part of pulp area that was filled with medical material. By applying linear regression on each set, mean of obtained slopes from all linear regression can be a good representative of rotation degree. Fig.5. explains briefly the algorithm of calculating rotation degree.

```
[max₁, max₂, ..., maxₙ]=maximum points of row (round (h/3))
Index=1;
Allocate n sets;
Set₁{index} =max₁, Set₂{index}=max₂, ..., Setₙ{index}=maxₙ
For i=round (h/3) +1 to round (2h/3)+1
  For set k=1 to n
    Find nearest max points in row (i) to Setₖ{index}
    Index=index+1
    Setₖ{index}=nearest max point position
  End
End
Apply linear regression to each set
[m₁, m₂,..., mₙ]=slopes of linear regression in sets
[deg₁, deg₂,..., degₙ]=arctan [m₁, m₂, ... mₙ]
Rotation degree=mean [deg₁, deg₂,..., degₙ]
```

Figure 5.  calculating of rotation degree algorithm

We have two ways to uses the result of above algorithm. One solution is to rotate the calculated line obtained from integral projection, another way we used is to rotate the image and do the integral projection again. Fig.6(a) illustrates the first way while Fig.6(b) express second solution. This algorithm can be iterative and will be stopped when there is no tangible change in the mean of obtained slopes.

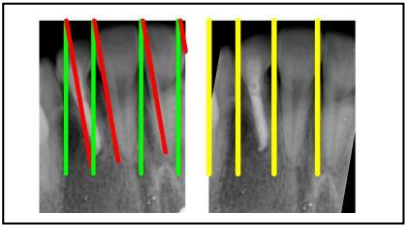

Figure 6.  (a) first solution. (b) second solution

## IV. EXPERIMENTAL RESULT

We can use our result as inputs to dental identification methods to evaluate the performance of segmentation algorithm, which is beyond our scope. In [11], said *et al* developed a metric-based object counting for the empirical evaluation of image segmentation. Fig.7 demonstrates experimental result of proposed method. We use Fig.7 to determine metrics of optimality, sub-optimality, and failure. Every cell $P_{ji}$ is the quantity of films having $i$ teeth but $j(j≤i)$ are segmented successfully;

$$F_i = \sum_{j=0}^{i} p_{ij} \qquad (3)$$

The performance metrics are defined as follows [11]:

$$optimality = 100 \frac{\sum_{i=1}^{N} p_{ii}F_i}{\sum_{i=1}^{N} F_i^2} \qquad (4)$$

$$failure = 100 \frac{\sum_{i=1}^{N} p_{0i}F_i}{\sum_{i=1}^{N} F_i^2} \qquad (5)$$

$$sub-optimality(m^{th}) = 100 \frac{\sum_{i=1}^{N-m} p_{i(i+m)}F_{i+m}}{\sum_{i=1}^{N} F_i^2} \qquad (6)$$

|   | 1 | 2 | 3 | 4 | 5 |
|---|---|---|---|---|---|
| 5 |   |   |   |   | 5 |
| 4 |   |   |   | 16 | 2 |
| 3 |   |   | 14 | 5 | 1 |
| 2 |   | 5 | 2 | 1 | 0 |
| 1 | 0 | 0 | 0 | 0 | 0 |
| 0 | 0 | 0 | 0 | 0 | 0 |
| # Scenes used in testing | 0 | 5 | 16 | 22 | 8 |

(Total # of correctly detected objects in scene vs Total # of object in scene)

Figure 7.  Testing result of proposed algorithm with Periapical radiograph

This proposed algorithm involves straight forward operations that can be operated in polynomial time, so there is no discussion about the time complexity. TABLE I represents the performance listing of proposed method.

TABLE I.  THE PERFORMANCE LISTING OF PROPOSED METHOD

| Opt | 1ˢᵗ | 2ⁿᵈ | 3ʳᵈ | Fail |
|---|---|---|---|---|
| 77.32 | 19.06 | 3.62 | 0 | 0 |

The squared mask size used in pre-processing step must be in logical range. Mask with large size makes image blurred and corrupts images while small mask has no visible change on images. Here, we use a 15*15 the squared mask. After that wiener filter was done on mean images. We apply a 2-D adaptive noise removal filtering which uses a pixel-wise adaptive wiener method base on statistical estimation from a local neighborhood of size m-by-n pixels to estimate local image mean and standard deviation. We consider a 10-by-10 neighborhood for using wiener filter. Finally we pass the obtained image through Gaussian filter.

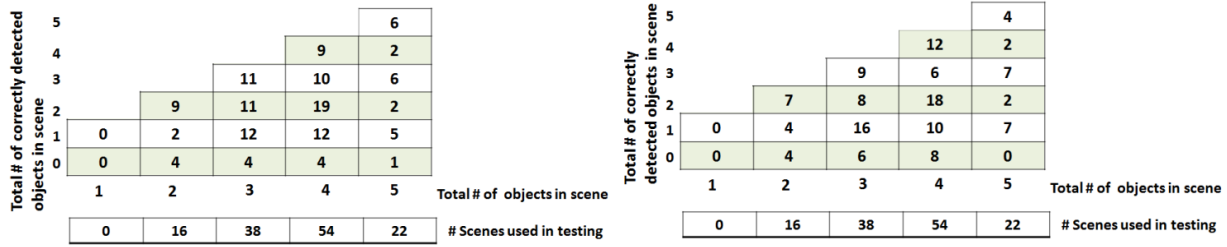

Figure 8. results of said et al segmentation algorithm with Periapical view. (b) result of said *et al* segmentation algorithm with enhanced Periapical view

The dataset we used contains 51 Periapical dental x-ray films. Our data are taken from school of dentistry at Shiraz University of medical science. Images in our database are not equal in size. Number of pixels in their widths is in the range of [450 650] and number of height pixels are in the range of [600 850].

We have assumed that difference between gaps must be more than a threshold; if the difference between two valleys is fewer than threshold, we select the valley which has less value in integral projection histogram. The maximum number of in our images dataset is five; we set the threshold to width of image divided by 6.

Fig.8 (a) shows testing result of said *et al* [11] segmentation algorithm with Periapical view and Fig.8 (b) is the result of algorithm with contrast-stretched Periapical view. Because our segmentation method is developed for segmenting Periapical radiographs, we compared it with segmentation methods that applied on Periapical radiograph. Fig.9. shows a graphical comparison using metric in [11] between proposed algorithm and algorithm is optimality presented at [11]. As it is shown proposed algorithm has better performance than the others.

Fig.10 illustrates some examples of segmented teeth. Blue lines express boundaries between teeth. Although the images quality in our database is poor, proposed method performance in good on test images.

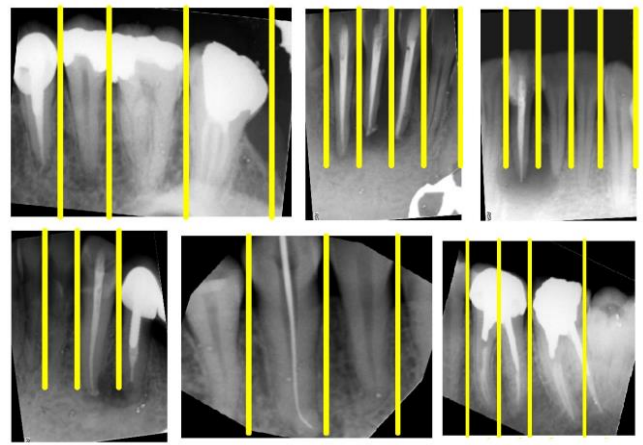

Figure 10. some example result of algorithm on test images.

## V. CONCLUSION AND FUTURE WORK

We use the property of root-canalled images in order to develop Periapical segmentation algorithm. This method can be used for any Periapical radiograph that contains at least one root-canalled image. Since most of people have been under endodontic treatment during their life, this method can be used in most of Periapical cases. Because the objective of Periapical view is to capture the tip of the root on the film, we are going to use this algorithm to develop a root apex detection algorithm for these kinds of images.

## REFERENCES

[1] Periodontal disease. CariFree. [Online] 2012. http://carifree.com/patient/cavities-a-z/periodontal-disease.html.

[2] The radiography information resource for patients. radiologyinfo.[Online]2012. http://www.radiologyinfo.org/en/info.cfm?pg=bonerad.

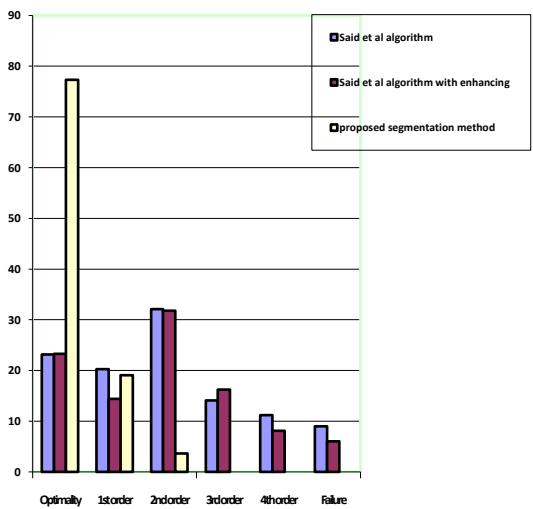

Figure 9. performance comparison between Periapical segmentation algorithms